\documentclass{article}

\usepackage{PRIMEarxiv}

\usepackage[utf8]{inputenc} 
\usepackage[T1]{fontenc}    
\usepackage{hyperref}       
\usepackage{url}            
\usepackage{booktabs}       
\usepackage{amsfonts}       
\usepackage{nicefrac}       
\usepackage{microtype}      
\usepackage{lipsum}
\usepackage{fancyhdr}       
\usepackage{graphicx}       
\graphicspath{{media/}}     
\usepackage{multicol}
\usepackage{amssymb}

\usepackage{mathtools}
\usepackage{algorithm}
\usepackage[noend]{algpseudocode}

\usepackage{enumitem}
\newlist{questions}{enumerate}{2}
\setlist[questions,1]{label=\textbf{RQ\arabic*.},ref=RQ\arabic*,leftmargin=*}
\setlist[questions,2]{label=(\alph*),ref=\thequestionsi(\alph*), leftmargin=*}

\usepackage{multirow}
\usepackage{graphicx}
\usepackage[table,xcdraw, dvipsnames]{xcolor}

\usepackage[normalem]{ulem}

\usepackage{tikz}

\usepackage{hyperref}

\usepackage{upgreek}

\usepackage{booktabs, multirow} 
\usepackage{soul}
\usepackage[table]{xcolor} 

\usepackage{capt-of}
\usepackage{tabu}
\usepackage{scalerel}
\usepackage{tikz}

\usepackage{blindtext}
\usepackage{geometry}
\usepackage{tikz}
\newcommand*\circled[1]{\tikz[baseline=(char.base)]{
            \node[shape=circle,draw,inner sep=2pt] (char) {#1};}}

\usepackage{wasysym}

\usepackage{caption}

\usepackage{comment}

\title{Generating Feasible and Plausible Counterfactual Explanations for Outcome Prediction of Business Processes}

\author{
  Alexander Stevens \\
  Research Centre for Information Systems Engineering (LIRIS) \\
  KU Leuven \\
  Leuven, Belgium\\
  \texttt{alexander.stevens@kuleuven.be}
  \And
  Chun Ouyang \\
   School of Information Systems \\
   Queensland University of Technology \\
  Brisbane, Australia
  \And
  Johannes De Smedt \\
  Research Centre for Information Systems Engineering (LIRIS) \\
  KU Leuven \\
  Leuven, Belgium
  \And
  Catarina Moreira \\
  Human Technology Institute \\
  University of Technology Sydney \\
  Sydney, Australia
}

\begin{document}
\maketitle

\begin{abstract}
In recent years, various machine and deep learning architectures have been successfully introduced to the field of predictive process analytics. Nevertheless, the inherent opacity of these algorithms poses a significant challenge for human decision-makers, hindering their ability to understand the reasoning behind the predictions. This growing concern has sparked the introduction of counterfactual explanations, designed as human-understandable ‘what if’ scenarios, to provide clearer insights into the decision-making process behind undesirable predictions. The generation of counterfactual explanations, however, encounters specific challenges when dealing with the sequential nature of the (business) process cases typically used in predictive process analytics. Our paper tackles this challenge by introducing a data-driven approach, REVISED$^{+}$, to generate more feasible and plausible counterfactual explanations. First, we restrict the counterfactual algorithm to generate counterfactuals that lie within a high-density region of the process data, ensuring that the proposed counterfactuals are realistic and feasible within the observed process data distribution. Additionally, we ensure plausibility by learning sequential patterns between the activities in the process cases, utilising Declare language templates. Finally, we evaluate the properties that define the validity of counterfactuals. 
\end{abstract}


\section{Introduction}\label{sec: introduction}
Over the past decade, the Business Process Management (BPM) domain has witnessed a strong uptake of data-driven process analyses. Many deep learning models have been successfully introduced to the field of Predictive Process Analytics (PPA), a subfield of BPM, where executed cases of processes within an information system are used \cite{van2016data}. For example, in a loan application process, every (process) case records the occurrence of sequentially ordered activities, starting from when the application is submitted by the customer up until the application form is closed. The focus of PPA is on identifying process-related trends such as impeding bottlenecks (e.g., how long will it take to process my loan application?), whether particular activities will occur in the future (e.g., will I receive an offer?) or predicting the outcome of the process (e.g., will my loan application be accepted or not?). In the latter field, the challenge of the predictive models is to predict the future state from observed data - an \emph{event} we aim to predict but has not occurred at the time of the inquiry. 

Despite the performance improvement achieved through the deployment of sophisticated models, their large parametric space - encompassing hundreds of layers and parameters - has resulted in a lack of transparency or \emph{black-boxness}~\cite{arrieta2020explainable}. This opacity poses a challenge, as users find it difficult to comprehend the reasons behind specific predictions, prompting the need for Explainable Artificial Intelligence (XAI) \cite{vilone2021notions}. In light of the AI transparency regulations described in the General Data Protection Regulation (GPDR) and the proposed European Union regulation for Artificial Intelligence, the AI Act, counterfactual explanations are seen as one of the most promising methods for providing human-understandable explanations \cite{wachter2017counterfactual, moreira2022benchmarking, guidotti2019factual, panigutti2023role}. While factual explanations help in understanding the relevance of each feature to the model prediction, counterfactual explanations, in the form of scenarios, provide users with insights about what changes in features lead to another prediction.\footnote{Note that we use \emph{counterfactual} in the context of XAI as initiated by \cite{wachter2017counterfactual}, as we imply an underlying machine or deep learning model and not a causal model.} This is also in contrast with the field of \emph{prescriptive} process monitoring, where the goal is to recommend a specific intervention \cite{donadello2023outcome, dasht2021prescriptive}. 

Moreover, if a predictive model predicts that a certain process instance is likely to deviate from the expected path (and resulting outcome), a counterfactual explanation might highlight the specific activities or events that, if different, could have led to another outcome. Therefore, we are specifically interested in cases with negative outcomes or process deviations, such as rejected loan applications, ICU admissions or equipment failures. The focus is on understanding the factors influencing the prediction, helping users comprehend the decision-making process of the model. Counterfactual scenarios are highly relevant in the field of PPA \cite{buliga2023counterfactuals} as this can help to reason about cases predicted to deviate from the desired path, as they emphasise the necessary changes in the activities in the cases required for the model to predict a different outcome. 

Unfortunately, the sequential structure of the data used for predictive process analytics, i.e. business process cases, is inherently different from the data structure of other types of data such as tabular data, time series etc. For example, time series data is a sequence of observations recorded at regular time intervals, where each data point is associated with a specific timestamp. business process data, on the other hand, captures a series of discrete events or activities in chronological order. These events are often associated with specific cases or processes and are not necessarily tied to regular time intervals. This means that the existing generation methods are not fit for purpose \cite{brunk2021cause}. Addressing the complexities of generating counterfactual explanations in PPA, this paper investigates the following research questions:\\

\begin{questions}
  \item \emph{What properties define the validity of counterfactual explanations in the field of predictive process analytics?}\\
 \item \emph{How should we generate counterfactual explanation methods that are tailored for predictive process analytics?}\\
\end{questions}

To answer the first research question \emph{RQ1.}, we need to extract the key properties that define the validity of counterfactuals in the field of counterfactual XAI (see \cite{chou2022counterfactuals} and \cite{verma2020counterfactual} for more information). Next, it is required that we translate the identified properties and their relevant evaluation metrics to make them suitable for a process-based analysis. This step is essential as blindly relying on quantitative evaluation metrics without tailoring them to the specific use case may lead to implausible or infeasible explanations \cite{moreira2022benchmarking}.

To address the second research question \emph{RQ2.}, we focus on the design of our novel approach for obtaining \emph{feasible} and \emph{plausible} counterfactual explanations for process outcome predictions. First, we confine the counterfactual algorithm to only generating counterfactuals that lie within a high-density region of the process data, ensuring that the proposed counterfactuals are realistic and \emph{feasible} within the context of the observed data distribution. Second, we learn sequential patterns between the activities in the cases using Declare language templates. We tailor our counterfactual generation approach to include sequential patterns that are 100\% supported by the cases in the dataset, as well as \emph{label-specific} sequential patterns that are observed in 100\% of cases with the desired outcome. Both steps ensure that the generated counterfactuals are aligned with the historical cases. Finally, we introduce an assessment framework to evaluate the overall validity of our counterfactual generation algorithm.

The rest of the paper is organised as follows. First, we describe the background and preliminaries in Section~\ref{sec: background}. Next, we define our counterfactual explanation algorithm for process outcome prediction in Section~\ref{sec: counterfactuals}. The experiment and implementation details can be found in Section~\ref{sec: experiment}. In the next Section~\ref{sec: discussion}, we evaluate our counterfactual generation algorithm on nine different event logs and demonstrate how the introduced algorithm is successful in enhancing both the quantity and quality of the generated counterfactuals. Next, we also provide an example of a counterfactual to show the usefulness of our approach on a real-life dataset. A review of the literature regarding (counterfactual) explainability in predictive process analytics is given in Section~\ref{sec: relatedwork}. Finally, the results, insights, and implications of this work are concluded in Section~\ref{sec: conclusion}. 

\section{Background}\label{sec: background}
\subsection{Business Process Cases}
The process data used in this research domain is commonly referred to as \emph{event logs}, as the occurrence of an activity in a process (case) is referred to as an `event'. In the context of process execution traces, an event log $L$ consists of events grouped per case.  Each event $e$ from the event universe $\xi$ is represented as a tuple $e = (c, a, t)$, where $c \in C$ denotes the case ID, $a \in A$, and $t \in \mathbb{R}$ denotes the timestamp. Consequently, a trace is a sequence of $n$ events $\sigma_c = [e_{1},e_{2},\dots,e_{i},\dots,e_{n}]$ such that $c$ is the case ID and $\forall i, j \in [1..n]$, $e_i .c = e_j .c$. For example, event $i$ in trace $j$ is denoted as $e_{i,j} = (c_{j}, a_{i,j},t_{i,j})$. The universe of all possible traces is denoted by $S$. Considering the varying lengths of these traces, let \( |\sigma_{c}| \) represent the length of any trace \(c\). The length of the longest trace is denoted as $|\sigma_{c}|_{max}$. 

To incrementally learn from the different stages of these traces, a prefix log denoted as $\mathbb{L}$ is often derived from the event log $L$. This prefix log encompasses all the prefixes originating from the full traces $\sigma$. A trace prefix for a given case $c$ with a length of $l$ is characterised as $\sigma_{c,l} = [e_{1},e_{2},\dots,e_{l}]$, where $l\leq |\sigma_c|$.  The set of activity types $|A|$ occurring in an event log is also referred to as the \emph{vocabulary} of the traces. Another often performed preprocessing step, i.e. trace cutting, involves limiting the prefix to a specific number of events. Moreover, trace cutting is implemented when the class label of the case relies on the presence of an activity; otherwise, the class label becomes known and irreversible \cite{teinemaa2019outcome}. A sequence encoding mechanism is necessary when working with machine or deep learning models and varying attribute amounts. For step-based models like recurrent neural networks, categorical dynamic attributes are transformed into a vector of continuous vectors using one-hot encoding, capturing the sequential nature of process data. The resulting vectors are denoted as $\sigma_{c} = (x_{1},x_{2},\dots,x_{m})$. Finally, the outcomes $y_{c}$ of the traces are dependent on the objectives of the process owner \cite{dumas2018fundamentals}. In this paper, we assume binary outcomes. The transformed log is utilised to train a predictive model $F$ that is able to predict the probability that the given trace belongs to the positive class. This is denoted as $\hat y_i = F(x_{i,1},\dots,x_{i,m})$. Therefore, the value $m$ can be as large as the product of the length of the longest trace and the number of activity labels, i.e., $ m \leq |\sigma_{c}|_{max} \times |A|$.

\subsection{Properties for Counterfactual Explanations}
Rather than supporting the understanding of why the predictive model made the specific prediction \cite{chou2022counterfactuals}, counterfactual explanations help the user to understand what changes would be necessary to reach the desired outcome (prediction) when the original prediction is undesired. Therefore, counterfactuals are often defined as instances \emph{as close as possible} to the original instance, with the predicted label for a counterfactual instance being the flipped label of the original instance. In the context of PPA, a counterfactual trace is denoted as $\sigma^{CF}_{c}$. The predictive model $F$ is used to predict the probability of the outcome of the counterfactual $\sigma^{CF}_{c}$ and is denoted as $\hat y^{CF}_{c} = F(x^{CF}_{c,1}, \dots, x^{CF}_{c,m})$. While it is important to realise that not every counterfactual is useful, there exist defined properties that guide the generation of \emph{viable} counterfactuals \cite{chou2022counterfactuals, verma2020counterfactual} .\\

\textbf{Proximity}. This property describes how close the counterfactual is to the original data point. In \cite{moreira2022benchmarking} and \cite{verma2020counterfactual}, it is argued that counterfactuals should be close to the input data point to improve the usefulness of the explanation. For instance, the counterfactual explanation search algorithm of Wachter et al. \cite{wachter2017counterfactual} generates counterfactual explanations by minimising the distance between a data instance $x$ and a counterfactual candidate $x^{CF}$. In the context of PPA, this would allow for the exploration of what (minimal) changes in the original trace could lead to a more desirable result. However, making changes to the trace without e.g. taking the order of the activities into account may result in implausible or nonviable scenarios. Similarly, in Natural Language Processing (NLP), making uncontrolled changes to the text data can result in ungrammatical counterfactuals \cite{madaan2021generate}. This claim is supported by Moreira et al. \cite{moreira2022benchmarking}, stating that counterfactual algorithms based uniquely on minimising proximity do not always provide meaningful counterfactuals.\\

\textbf{Sparsity}. This property describes the number of changed features between the initial instance and the counterfactual instance. In \cite{mothilal2020explaining} and \cite{verma2020counterfactual}, it is stated that a counterfactual generation algorithm should aim to change as few attributes as possible, suggesting that shorter explanations tend to be more user-friendly and easier to understand. However, in the context of PPA, not taking into account the order of the activities while minimising the number of changes may result in implausible counterfactual scenarios.\\

\textbf{Diversity}. This property defines the diversity (i.e. variety) amongst the different counterfactual explanations by comparing the pairwise distances between each of the resulting counterfactuals. Some algorithms, such as Diverse Counterfactual Explanations (DICE) \cite{mothilal2020explaining}, support and optimise for having a diverse set of counterfactuals by adding a diversity term to the objective function. The argument is that multiple, diverse counterfactuals allow for more interpretable and human understandable explanations. \\
\begin{figure}[h]
\centering
\includegraphics[width=0.5\columnwidth]{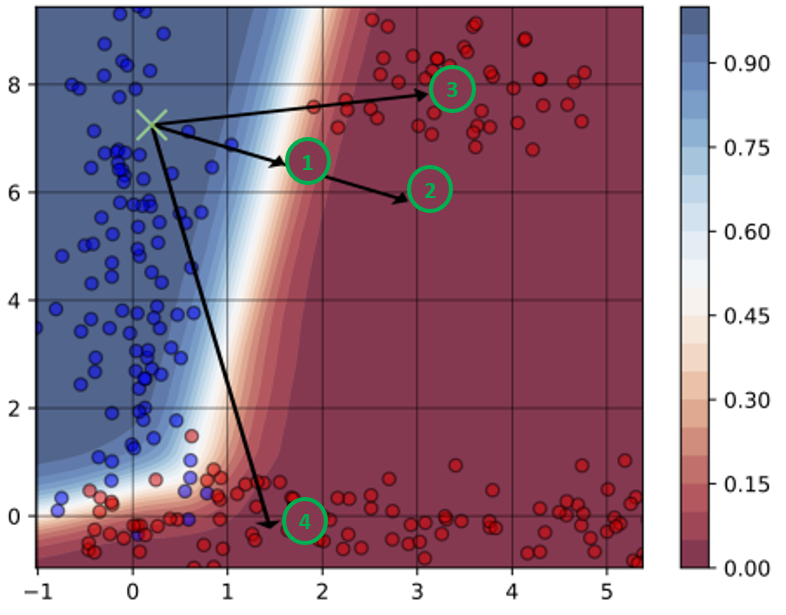}
\caption{The counterfactual instances \protect\circled{1}, \protect\circled{2}, \protect\circled{3} and \protect\circled{4}. Both \protect\circled{1} and \protect\circled{2} lie in low-density regions and are therefore considered \emph{infeasible}. The counterfactual \protect\circled{4} is preferred over \protect\circled{3} because there is a feasible path between the initial data point and the counterfactual. This figure is adapted from \cite{poyiadzi2020face}}.
\label{fig: FACE}
\end{figure}

\textbf{Plausibility}. This property defines how plausible or believable, actionable and reasonable a counterfactual is. Most studies in the literature highlight this as one of the important properties \cite{chou2022counterfactuals, moreira2022benchmarking, verma2020counterfactual}. This therefore also means that certain \emph{immutable} attributes such as gender or race should never be changed if plausibility needs to be assured. \cite{moreira2022benchmarking} states that having meaningful evaluation results hinges on guaranteeing plausibility in the internal mechanisms of the counterfactual generation process. In the case of the above-mentioned loan application process, this means that the application process must adhere to certain process requirements e.g. a client cannot apply twice for the same loan. \\

\textbf{Feasibility}. This property is introduced by \cite{poyiadzi2020face} and indicates whether the difference between the initial data point and the counterfactual is generated through feasible changes. Moreover, the authors argue that the closest counterfactual to a data instance does not necessarily lead to a feasible change in the attribute. e.g., low-skilled unsuccessful mortgage applicants are told to double their salary to have their loan accepted, which is not easy to realise, or even \emph{unrealistic} as counterfactual. Furthermore, in \cite{brown2021uncertainty}, it is argued that the uncertainty of their counterfactuals is high because these counterfactuals may lie outside the explored decision space. This is described as \emph{data manifold closeness} in \cite{verma2020counterfactual}, which states that a generated counterfactual should be realistic in the sense that it is near the training data and adheres to observed correlations among the attributes. Consider the four possible counterfactuals presented In Figure \ref{fig: FACE}. All these counterfactuals are viable as they lie across the decision boundary. Counterfactual A is the closest to the initial data point and is the counterfactual that would be preferred if we purely optimise for \emph{proximity} \cite{wachter2017counterfactual}. However, as both counterfactuals A and B lie in a low-density region, they are deemed infeasible as there are almost no precedents of similar instances in the data. For counterfactual C, there is no feasible path between the initial data instance and the suggested counterfactual therefore making the latter infeasible as well \cite{poyiadzi2020face}.

\subsection{Manifold Learning}

Previous approaches have already dealt with the adherence to the data manifold in many different ways, e.g. by constraining the counterfactual explanations to the underlying distribution \cite{joshi2019towards, pawelczyk2020learning, DBLP:conf/hicss/ForsterHKK21, artelt2021efficient, schut2021generating, dhurandhar2018explanations,van2021interpretable, van2021conditional, artelt2020convex}. The manifold hypothesis \cite{connor2021variational} suggests that high-dimensional data can be effectively represented on a low-dimensional manifold. Therefore, the above approaches use generative models such as variational autoencoders (VAEs) or Generative Adversarial Networks (GANs) to learn mappings from low-dimensional latent vectors to high-dimensional data while encouraging these latent variables to adhere to an assumed prior distribution such as a Gaussian distribution. From now on, we will refer to this as \emph{manifold learning}.

Manifold learning in PPA has already been introduced in  \cite{stevens2023manifold}, where they trained VAEs on the distribution of the data. In contrast to a simple (non-variational) autoencoder, where the primary objective is to reconstruct the input $(x)$ through the encoding process of an encoder $e$ to generate a latent representation $z = e(x)$, variational autoencoders approach this as a probabilistic challenge. The objective of a VAE is to maximise the Evidence Lower Bound (ELBO) which equates to striking an optimal balance between the reconstruction loss (\emph{"how well can we reconstruct the initial data point?"}) and the Kullback-Leibler (KL) divergence (\emph{"how close is the chosen prior distribution to the learned latent variable distribution?"}). In the context of training VAEs, the prior distribution $P$ is often chosen as the standard normal distribution, and the learned latent variable distribution $Q$ is the one approximated by the encoder in the VAE.

The reconstruction loss is typically measured by the Negative Loss-Likelihood (NLL) loss:

\begin{equation}
\begin{split}
L(\theta)_{NLL} = -\sum_{\sigma_c \in S} & y_{\sigma_c} \cdot \log(\hat{y}_{\theta,\sigma_c}) \\
& + (1 - y_{\sigma_c}) \cdot \log(1 - \hat{y}_{\theta,\sigma_c})
\end{split}
\label{eq:NL}
\end{equation}

and the KL divergence is the sum taken over all components of the distributions (i.e. the number of latent variables in the latent space, $r$, and $r \leq p$):

\begin{center}
\begin{align}
L_{KL}(P || Q) = \sum_{\sigma_c \in S} P(\sigma_c) \cdot \log\left(\frac{P(\sigma_c)}{Q(\sigma_c)}\right) \label{eq:DKL}
\end{align}
\end{center}

\subsection{The Declare Language}

\begin{table*}[ht]
    \centering
    \captionsetup{skip=8pt}  
    \caption{An Overview of Declare templates with their corresponding LTL formula. This table is adapted from \cite{de2019mining}.}

    \label{tab:declareconstraints}
    \resizebox{0.8\textwidth}{!}{%
        \begin{tabular}{|l|l|l|}
            \hline
            \textbf{Type} & \textbf{Template} & \textbf{LTL Formula \cite{pesic2008constraint}} \\ \hline
            \multirow{5}{*}{Unary} &
            Existence(a,n) &
            $\Diamond(a\wedge\bigcirc(existence(n-1,a)))$ \\
            &
            absence(a,n) &
            $\neg existence(n,a)$ \\
            &
            Exactly(a,n) &
            $existence(n,a) \wedge absence(n+1,a)$ \\
            &
            Init(a) &
            $a$ \\
            &
            Last(a) &
            $\Box(a\implies\neg X\neg a)$ \\ \hline
            \multirow{1}{*}{Unordered} &
            Co-existence(a,b) &
            $\Diamond a\Longleftrightarrow\Diamond b$ \\ 
            \hline
            \multirow{4}{*}{Simple Ordered} &
            Response(a,b) &
            $\Box(a\implies\Diamond b)$  \\
            &
            Precedence(a,b) &
            $(\neg b\, Ua)\vee\Box(\neg b)$ \\
            &
            Succession(a,b) &
            $response(a,b)\wedge precedence(a,b)$ \\
            &
            Not succession(a,b) &
            $\Box(a\implies\neg(\Diamond b))$ \\ \hline
            \multirow{6}{*}{Ordered}
            &
            alternate response(a,b) &
            $\Box(a\implies\bigcirc(\neg a\, U\, b))$ \\ 
            &
            alternate precedence(a,b) & $precedence(a,b)\wedge\Box(b\implies\bigcirc(precedence(a,b))$ \\ 
            &
            alternate succession(a,b) &
            $altresponse(a,b)\wedge precedence(a,b)$ \\ 
            &
            Chain response(a,b) &
            $\Box(a\implies\bigcirc b)$ \\
            &
            Chain precedence(a,b) &
            $\Box(\bigcirc b\implies a)$ \\
            &
            Chain succession(a,b) &
            $\Box(a\iff\bigcirc b)$ \\ \hline
        \end{tabular}%
    }
\vspace{1pt} 
\end{table*}

Declare \cite{pesic2007declare} is a language that is often used for describing declarative process models based on Linear Temporal Logic (LTL) on Finite traces, i.e. $LTL_{f}$ \cite{de2013linear}. These process models provide a high-level specification of the desired behaviour in a process and capture the temporal order and dependencies of activities within a process. The language uses patterns which cover a wide range of behavioural properties such as the appearance of an activity in a trace, the eventual ordering between two activities, and so on. These patterns are often used in the field of process mining to identify sequential or behavioural characteristics of processes \cite{di2013two}. Moreover, we can observe the \emph{support} of a certain pattern within the process data by calculating the proportion of cases that have the pattern. For example, a support value of 50\% implies that at least half of the observed cases must have that certain pattern. In this paper, we only use a support level of 100\% as we solely focus on the most frequent patterns. 

The use of these patterns makes Declare excel at capturing a vectorial representation of process model behaviour, contrary to other behavioural languages such as procedural models such as Petri nets. This makes them ideal in various data-driven models such as predictive process models.
The interesting Behavioural Constraint Miner (iBCM) \cite{de2019mining} can derive sequential patterns (or constraints) that are based on these high-level Declare language templates \cite{donadello2023outcome, di2013two}. 
The set of templates that are used by the iBCM algorithm is listed in Table \ref{tab:declareconstraints} and divided into four categories: \emph{unary}, \emph{unordered}, \emph{simple ordered} and \emph{ordered}. The unary constraints either focus on the position (first/last) or the cardinality (existence/absence/exactly) of the activity. The unordered constraints do not take into account the order of the activities, i.e. \emph{co-existence(A, B)} expresses that the presence of $A$ requires the presence of $B$ without specifying their relative location. Next, the simple ordered constraints form the base for all the rest of the binary constraints. \emph{Precedence(A,B)} defines that $B$ only occurs if it is preceded by $A$, while \emph{response(A,B)} requires that whenever $A$ happens, $B$ has to happen eventually afterwards. Furthermore, \emph{succession(A, B)} states that both \emph{precedence(A, B)} and \emph{response(A, B)} should hold. The constraint \emph{not succession} is to express that the constraint \emph{succession(A, B)} did not occur. The alternating-ordered constraints introduce the concept of alternating patterns in the occurrence of activities, by stating that you cannot have two $As$ (or $Bs$) in a row before an $A$ (or $B$). The chain-ordered patterns focus on the relationship with the last occurrence of one of the activities, stating that no other activity can occur between $A$ and $B$. For instance, the presence of \emph{chain precedence(A, B)} implies the presence of \emph{alternate precedence(A, B)}, and \emph{precedence(A, B)}, as well as \emph{responded existence(B, A)}. This allows for expressing the presence of certain groups of activities, their ordering, and also the repeating (alternation) and local (chain) behaviour. 
\section{Counterfactual Explanation Generation in Predictive Process Analytics}\label{sec: counterfactuals}

In this section, we describe our strategy for creating viable counterfactual explanations in predictive process analytics (PPA). Our approach prioritises key properties like plausibility and feasibility, particularly concerning process-based analysis, by considering the control flow constraints. To this end, it is required that we elaborate on how these properties are translated to the field of PPA.

\subsection{Counterfactual Properties in the context of PPA}

\textbf{Plausibility in PPA.} In \cite{moreira2022benchmarking}, plausibility defines that immutable features (such as gender or age) should never be changed. Therefore, we argue in this paper that plausibility must be treated as a ``hard requirement", indicating that counterfactuals must strongly adhere to certain predefined criteria to be deemed plausible or believable. Nonetheless, plausibility in the context of a sequential data structure such as business processes has a different meaning compared to tabular data, and the notion of \emph{immutable features} is hence different. Therefore, we claim that a \emph{plausible counterfactual trace} should satisfy both the process trace constraints and the label-specific trace constraints to be considered as \emph{plausible}. The plausible rate describes the number of counterfactuals without Declare constraint violations (in \%). Moreover, we can calculate the number of violations based on the Declare Constraints (DC) as follows:

\begin{equation}\label{eq:DC}
\begin{aligned}
L(\theta)_{DC} &= \frac{1}{N} \sum_{\sigma_c \in S} w_j \phi_{\sigma_c,j} \\
&\quad \text{where} \quad \phi_{\sigma_c,j} = 
\begin{cases}
    1, & \text{if constraint $j$ is violated} \\
    0, & \text{else}
\end{cases}
\end{aligned}
\end{equation}
and $w_{j}$ is the importance per constraint.\\

Based on the domain knowledge of the stakeholders, it is possible to choose different importance weights per constraint. In this paper, we assume equal importance weights i.e. $w_{1} = w_{2} \dots = w_{N} = 1$. The revised loss function of the VAE is now:

\begin{equation}
\begin{split}
L_{VAE} = \lambda_{NLL} \cdot L_{NLL} + \lambda_{KL} \cdot L_{KL}(P || Q) \\ + \lambda_{DTC} \cdot L_{DTC}
\end{split}
\label{eq:VAEloss}
\end{equation}

The last term in this loss function penalises violations of Declare constraints we find in any of the traces, regardless of what label they belong to. These are referred to as Declare Trace Constraints (DTC). We must add these constraints when we use the variational autoencoder to learn the data distribution as we argue that the trace constraints are constraints that, as they are present in all the sequences, define \emph{how a trace should look like}.

\begin{figure}[]
\centering
\includegraphics[width=1\columnwidth]{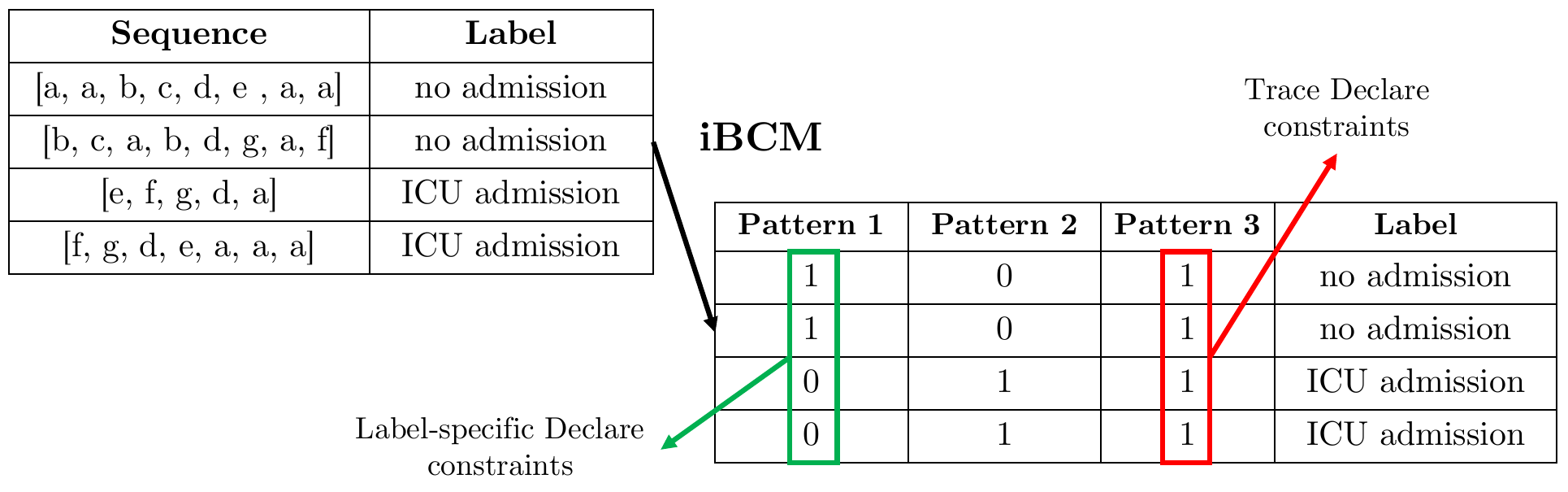}
\caption{A simplified example of how the iBCM algorithm extracts the different trace and label-specific constraints. This figure is adapted from \cite{deeva2022predicting}.}
\label{fig: declarefig}
\end{figure}

The Label-specific Declare Constraints (DLC), on the other hand, are the Declare constraints mined from the traces with the desired label, and we add these to the loss function of the counterfactual generation algorithm (see Section \ref{subsec: counterfactualgeneration}). The label-specific constraints are constraints that are only 100\% supported by a specific label and therefore define \emph{what a trace of a certain label should look like}. This is also visualised in Figure\ref{fig: declarefig}.\\

\textbf{Feasibility in PPA.} Feasibility, sometimes also referred to as actionability \cite{karimi2022survey}, is one of the most difficult but also most important properties. In \cite{chou2022counterfactuals}, it is argued that \emph{feasibility} implies \emph{plausibility}. However, here we argue that the following negative implication makes more sense in the field of PPA: a violation of \emph{plausibility} implies a violation of \emph{feasibility} as well. 

Feasibility is satisfied when the counterfactual algorithm suggests useful (and feasible) changes for the decision-maker to take action. Moreover, a counterfactual is a better candidate if it falls in a well-defined region of the decision boundary and corresponds to the point with the shortest path to the data instance. Another possibility to evaluate the feasibility of counterfactuals in PPA could be to calculate the remaining time of the counterfactual as a “cost” and see whether the counterfactual is feasible. Nonetheless, this approach might be more dependent on the quality of the remaining time prediction model than the quality of the counterfactual itself but would be a simple and intuitive assessment. Next, it would also be possible to calculate some sort of feasibility metric, such as done in \cite{polyvyanyy2020entropic}, where an entropic relevance measure to calculate the feasibility in the context of conformance checking is introduced. Another measure is the probability of a sequence as the product of the transition probability multiplied by the state emission probability for each step in the counterfactual sequence \cite{hundogan2023created}. However, as our counterfactual generation algorithm is already restricted to the data manifold, we abstract from these feasibility metrics.\\

\textbf{Proximity in PPA.} Proximity between the initial instance and counterfactual explanation is often measured with $L_{p}$ norms such as the $L_{1}$ norm, $L_{2}$ norm, and the $L_{\inf}$ norm \cite{moreira2022benchmarking}. In practice, most approaches use a weighted combination of multiple distance functions such as the elastic net regulariser ($L_{1}$/12 combination) \cite{zou2005regularization} of the Prototype algorithm \cite{van2021interpretable} or the $L_{0}$/$L_{2}$ combination of Growing Spheres algorithm \cite{laugel2018comparison}. From a process perspective, a distance metric such as the Earth Mover Distance (EMD) is often used as well \cite{peeperkorn2023global}. In this paper, we use the elastic net regulariser which is a weighted distance loss function of both the \emph{$L_{1}$ norm} and \emph{$L_{2}$ norm} to optimise for \emph{proximity}. 

\begin{center}
\begin{minipage}{0.9\linewidth}
The elastic net regulariser \cite{zou2005regularization} is defined as:
\begin{equation}\label{eq:elasticnet}
\text{$L_{dist}$} = \beta \cdot \lVert \delta \rVert_1 + \lVert \delta \rVert_2^2 = \beta \cdot \text{$L_{1}$} + \text{$L_{2}$}
\end{equation}
\end{minipage}%
\end{center}

where $\delta$ is the subtraction between the input data point and the counterfactual (i.e. input - counterfactual). $\beta$ is a scaling factor, and $\lVert \delta \rVert_1$ and $\lVert \delta \rVert_2^2$ are the $L_{1}$ and $L_{2}$ norms, respectively.\\

\textbf{Sparsity in PPA} Sparsity in the field of XAI is often measured with the use of the $L_{0}$ norm, by counting the number of nonzero elements between the difference of the initial and the counterfactual explanation. Another interesting measure in the field of PPA is the Damerau-Levenshtein (DL) distance \cite{damerau1964technique}. However, none of these metrics takes the order of the activities into account, which is an important aspect in the field of PPA. Therefore, we propose to use the Longest Common Prefix (LCP) \cite{karkkainen2009permuted} metric, which calculates the number of events (starting from the first event) to which the counterfactual is still similar compared to the initial trace.\\

\textbf{Diversity in PPA.} Many algorithms only support the generation of a single counterfactual, with Wachter et al. \cite{wachter2017counterfactual} highlighting the fact that providing a diverse set of counterfactual explanations is more informative than providing the "closest possible" counterfactual according to a distance metric. The property is further elaborated on in \cite{russell2019efficient}. The idea is to generate a set of diverse counterfactual explanations for the same sequence. For example, in DICE \cite{mothilal2020explaining}, a diversity term is added to the loss function to optimise for diversity.

Although our introduced algorithm can generate multiple counterfactuals, our optimisation objective is not optimised for this as we perform iterative gradient steps to walk through the latent space.\\

\textbf{Other properties in PPA.}
Other interesting evaluation measures are e.g. \textit{y-NN} \cite{pawelczyk2021carla}, which is the fraction of nearest neighbours with a similar label compared to the counterfactual. Here, a value of 1 indicates that all the nearest neighbours have the same label as the predicted outcome of the counterfactual. This is related to the connectedness \cite{pawelczyk2020learning} of a generated counterfactual, with the reasoning here that counterfactuals that are connected to few correctly classified observations are deemed as “harder to predict”. Finally, the success rate describes the number of counterfactuals generated per total number of iterations (in \%) and $|$CF$|$ is the length of the counterfactual.

\subsection{Counterfactual Generation Algorithm}\label{subsec: counterfactualgeneration}

To this end, we use the existing REVISE counterfactual algorithm \cite{joshi2019towards} and perform multiple adaptations in multiple ways. First, we adapt the algorithm so that it can handle sequential data instead of only tabular data. This algorithm is referred to as REVISE$^+$. Second, we have adapted the algorithm for a process-based analysis by enhancing the optimisation objective of the variational autoencoder and the generation algorithm with trace process constraints and label-specific process constraints, i.e. REVISED$^+$. This ensures that our counterfactual algorithm optimises for \emph{plausibility}. Third, we use hinge loss for the class loss to allow for more counterfactual solutions as we only require the predicted probability $ \hat y^{CF}_{i} = F(x^{CF}_{1}, \dots, x^{CF}_{m})$ to be greater than a predefined threshold (e.g. 0.5). 

\begin{center}
\begin{minipage}{0.9\linewidth}
The hinge loss is given by:
\begin{equation}\label{eq:hingeloss}
\text{$L_{hinge}$} = \max(0, 1 - \zeta \cdot logit(\hat y_{\text{CF,i}})
\end{equation}
\end{minipage}%
\end{center}

where $\zeta = -1$ when $y = 0$ and $\zeta = 1$ when $y = 1$, and $logit(\hat y^{CF}_{i})$ is the unscaled output from the model. Fourth, we use the elastic net regulariser instead of a single distance metric. \\

The resulting loss function used for the counterfactual generation algorithm is:

\begin{equation}
\begin{split}
L(\sigma_{CF}) \gets + \lambda_{hinge} \cdot \text{$L_{hinge}$} 
    + \lambda_{dist} \cdot \text{$L_{dist}$} \\
    + \lambda_{DLC} \cdot L_{DLC} \label{eq:loss}
\end{split}
\end{equation}

\subsubsection{Step 1: Learning the Data Manifold of the Process Data}

The idea of the counterfactual generation algorithm is to perform gradient steps in the latent space of the learned data manifold and to generate counterfactuals from the latent sample of the initial data point. More specifically, learning the underlying representations of the process data with the use of LSTM-VAEs, allows us to generate \emph{actionable} counterfactuals. The manifold-restricted approach ensures that the counterfactual instance still exists in a high-density area. 

\subsubsection{Step 2: Generating the Counterfactuals}

The counterfactual generation method (pseudocode is given in Algorithm \ref{alg: counterfactual}) begins by initialising a latent sample, denoted as $z$, using the VAE encoder $E(x)$. This latent sample serves as the starting point for generating counterfactuals (line 2). Within an iterative loop, the algorithm generates a counterfactual trace, $\sigma^{CF}$, by decoding the current latent sample using the VAE decoder $D(x)$ (line 4). The loss function, $L(\sigma^{CF})$, comprises three components: a hinge loss punishing for high probabilities for the undesired label, a distance loss encouraging closeness to the original trace, and the DLC loss punishing for label-specific process constraint violations (line 6-8).
\begin{center}
\scalebox{0.8}{
\begin{minipage}{1.2\linewidth}
\begin{algorithm}[H]
\caption{Counterfactual Generation Algorithm for Feasible and Plausible Explanations}\label{alg: counterfactual}
\begin{algorithmic}[1]
\Require $x$, VAE encoder $E(x)$, VAE Decoder $D(x)$, $\lambda_{hinge}$, $\lambda_{dist}$, $\lambda_{DLC}$, threshold $p$, $List^{CF}$
\State $z \gets E(\sigma_{c})$ \Comment{Initialize latent sample for a trace}
\Repeat \Comment{repeat until stopping criteria}
    \State $\sigma^{CF} \gets D(z)$ \Comment{Initialize counterfactual}
    \State $L(\sigma^{CF}) \gets + \lambda_{hinge} \cdot \text{$L_{hinge}$} 
    + \lambda_{dist} \cdot \text{$L_{dist}$}
    + \lambda_{DLC} \cdot L_{DLC}$ \Comment{loss of $\sigma^{CF}$}
    \State $z \gets z - \alpha \cdot (\lambda_{hinge} \cdot \nabla L_{\text{hinge}} + \lambda_{dist} \cdot \nabla L_{\text{dist}} + \lambda_{DLC} \cdot \nabla L_{\text{DLC}})$ \Comment{gradient descent}
    \If{$F(\sigma^{CF}) > p$ and $L_{DLC} = 0$} \State $List^{CF}.\text{append}(\sigma^{CF})$ \Comment{Save $\sigma^{CF}$ to final CF list if it is a viable counterfactual}
    \EndIf
\Until{stopping criteria}   \Comment{maximum number of iterations}
\State \Return $List^{CF}$   \Comment{list of viable counterfactuals}
\end{algorithmic}
\end{algorithm}
\end{minipage}%
}
\end{center}

Gradient descent is then employed to iteratively update the latent sample $z$, moving it towards regions in the latent space that generate counterfactuals with reduced loss (line 10). The counterfactual is added to the final counterfactual list $List^{CF}$ (line 14) if two conditions are met: (1) the predicted probability exceeds a predefined threshold ($F(\sigma^{CF}) > p$, line 12) and (2) no violations of label-specific process constraints ($L_{DLC} = 0$, line 12). The iterative process continues until predefined stopping criteria are met (line 16), and the algorithm ultimately returns the list $List^{CF}$ (line 18).

\begin{table*}[]
\centering
\caption{An overview of the adaptations made compared to the original counterfactual generation algorithm REVISE \cite{joshi2019towards}, the adapted algorithm that works with the sequential structure of process data (REVISE$^{+}$) and the introduced method REVISED$^{+}$ that incorporates the Declare patterns in both the data manifold (TDC constraints) and in the counterfactual generation algorithm (the LDC constraints).}
\label{tab: REVISED}
\resizebox{\textwidth}{!}{%
\begin{tabular}{
>{\columncolor[HTML]{FFFFFF}}r 
>{\columncolor[HTML]{FFFFFF}}r |
>{\columncolor[HTML]{FFFFFF}}c |
>{\columncolor[HTML]{FFFFFF}}c |
>{\columncolor[HTML]{FFFFFF}}c }
\hline
\multicolumn{2}{c|}{\cellcolor[HTML]{FFFFFF}} &
  \cellcolor[HTML]{FFFFFF} &
  \cellcolor[HTML]{FFFFFF} &
  \cellcolor[HTML]{FFFFFF} \\
\multicolumn{2}{c|}{\cellcolor[HTML]{FFFFFF}} &
  \cellcolor[HTML]{FFFFFF} &
  \cellcolor[HTML]{FFFFFF} &
  \cellcolor[HTML]{FFFFFF} \\
\multicolumn{2}{c|}{\multirow{-3}{*}{\cellcolor[HTML]{FFFFFF}\textbf{}}} &
  \multirow{-3}{*}{\cellcolor[HTML]{FFFFFF}\textbf{REVISE}} &
  \multirow{-3}{*}{\cellcolor[HTML]{FFFFFF}\textbf{REVISE$^{+}$}} &
  \multirow{-3}{*}{\cellcolor[HTML]{FFFFFF}\textbf{REVISED$^{+}$}} \\ \hline
\multicolumn{2}{r|}{\cellcolor[HTML]{FFFFFF}Data Format} &
  {\color[HTML]{9B9B9B} tabular data} &
  3D temporal data &
  3D temporal data \\
\multicolumn{2}{r|}{\cellcolor[HTML]{FFFFFF}CFs} &
  {\color[HTML]{9B9B9B} single} &
  multiple &
  multiple \\ \hline
\multicolumn{2}{r|}{\cellcolor[HTML]{FFFFFF}Manifold Learning} &
  {\color[HTML]{9B9B9B} VAE} &
  LSTM VAE &
  LSTM VAE with Declare constraints \\
\multicolumn{2}{r|}{\cellcolor[HTML]{FFFFFF}VAE Loss Function} &
  {\color[HTML]{9B9B9B} $L_{VAE} = \lambda_{BCE} \cdot L_{NLL} + \lambda_{KL} \cdot L_{KL}(P || Q)$} &
  $L_{VAE} = \lambda_{NLL} \cdot L_{NLL} + \lambda_{KL} \cdot L_{KL}(P || Q)$ &
  \begin{tabular}[c]{@{}c@{}}$L_{VAE} = \lambda_{NLL} \cdot L_{NLL} + \lambda_{KL} \cdot L_{KL}(P || Q)$\\ $+ \lambda_{DTC} \cdot L_{DTC}$\end{tabular} \\ \hline
\multicolumn{2}{r|}{\cellcolor[HTML]{FFFFFF}CF Loss Function} &
  {\color[HTML]{9B9B9B} $L(\sigma_{CF}) = \lambda_{BCE} \cdot L_{BCE} + \lambda_{dist} \cdot L_{dist}$} &
  $L(\sigma_{CF})  =  \lambda_{hinge} \cdot \ L_{hinge} + \lambda_{dist} \cdot L_{dist}$ &
  \begin{tabular}[c]{@{}c@{}}$L(\sigma_{CF})  = \lambda_{hinge} \cdot L_{hinge} + \lambda_{dist} \cdot L_{dist}$\\ $+ \lambda_{DLC} \cdot L_{DLC}$\end{tabular} \\ \hline
\cellcolor[HTML]{FFFFFF} &
  Proximity &
  {\color[HTML]{9B9B9B} $L_{1}, L_{2}, L_{\inf}$} &
  $L_{1}, L_{2}$, EMD $\cite{peeperkorn2023global}$ &
  $L_{1}, L_{2}$, EMD $\cite{peeperkorn2023global}$ \\
\cellcolor[HTML]{FFFFFF} &
  Sparsity &
  {\color[HTML]{9B9B9B} $L_{0}$, redundancy} &
  $L_{0}$, DL edit $\cite{damerau1964technique}$, LCP $\cite{karkkainen2009permuted}$ &
  $L_{0}$, DL edit $\cite{damerau1964technique}$, LCP $\cite{karkkainen2009permuted}$ \\
\cellcolor[HTML]{FFFFFF} &
  Plausibility &
  {\color[HTML]{9B9B9B} plausible ranges, immutable features} &
  TDC, LDC, plausible rate (\%) &
  TDC, LDC, plausible rate (\%) \\
\cellcolor[HTML]{FFFFFF} &
  Feasibility &
  {\color[HTML]{9B9B9B} restricted by VAE} &
  restricted by LSTM VAE &
  restricted by LSTM VAE with Declare constraints \\
\cellcolor[HTML]{FFFFFF} &
  Diversity &
  {\color[HTML]{9B9B9B} N/A} &
  $L_{\text{div.}} = \frac{1}{\sum_{i=1}^{n-1} \sum_{j=i+1}^{n} \text{EMD}(i, j)}$ &
  $L_{\text{div.}} = \frac{1}{\sum_{i=1}^{n-1} \sum_{j=i+1}^{n} \text{EMD}(i, j)}$ \\
\multirow{-6}{*}{\cellcolor[HTML]{FFFFFF}\begin{tabular}[c]{@{}r@{}}Evaluation \\ framework\end{tabular}} &
  Extra &
  {\color[HTML]{9B9B9B} Y-NN, success rate (\%), run time} &
  Y-NN, success rate (\%), $|$CF$|$ &
  Y-NN, success rate (\%), $|$CF$|$ \\ \hline
\end{tabular}%
}
\end{table*}
\section{Experimental Results}\label{sec: experiment}

In this section, the objective is to generate plausible and feasible counterfactuals for the cases with predicted negative outcomes in all nine event logs. To this end, we compare the counterfactual generation algorithms REVISE$^{+}$ and REVISED$^{+}$ (see Table \ref{tab: REVISED}) to evaluate whether the incorporation of Declare constraints in the loss functions of our counterfactual generation algorithm improves the validity of the counterfactuals. Table \ref{tab: results} provides a concise overview of the evaluation metrics that define the key properties defining counterfactual validity. Finally, we provide an example of a counterfactual for a specific case of the sepsis event log in Table \ref{tab: onefactual}.

\subsection{Event Logs}

The event logs used are \href{https://data.4tu.nl/articles/dataset/Sepsis_Cases_-_Event_Log/12707639}{sepsis cases}, \href{https://data.4tu.nl/articles/dataset/BPI_Challenge_2012/12689204}{BPIC2012} and \href{https://doi.org/10.4121/uuid:31a308ef-c844-48da-948c-305d167a0ec1}{BPIC2015}. The sepsis cases record information about patients who are admitted to a Dutch hospital with symptoms of a life-threatening sepsis condition. In this event log, every event, from the registration of the patient in the emergency room up until the final discharge from the hospital is recorded. The labelling in sepsis cases is based on whether the patient is eventually admitted to intensive care `(yes)` or not `(no)`. There are 85 deviant cases and 540 regular cases. The BPIC2012 event log contains information about the execution history of a loan application process in a Dutch Financial institution. For BPIC2012, the labelling is done based on whether the loan is accepted `(yes)` or not `(no)`. Finally, the BPIC2015 dataset consists of event logs from 5 Dutch municipalities, each describing their building permit application process. The labelling of each event log is based on the satisfaction of an LTL rule as defined in \cite{teinemaa2019outcome}.

\subsection{Implementation Design Choices}

First, a temporal train-test split is performed that ensures that the period of training data does not overlap with the period of the test data, while the events of the cases in the train data that did overlap with the test data are cut. Next, we remove the event (or activity) that contains information about the outcome. For example, if the outcome is to predict whether the patient will be admitted to the ICU (or not), we only keep the sequence up until this activity (to avoid leaking the outcome). Note that we only consider the full traces (and not trace prefixes) to make sure that we can learn the Declare constraints that describe the process behaviour of the cases. We perform trace cutting on the 90th quartile of trace lengths to prevent excessively long traces. This additionally ensures that our prediction model is not biased towards longer cases \cite{teinemaa2019outcome}.

Finally, we add an artificial End of Sequence (EoS) token so that the model can learn (and we can observe) what the end of the trace is.  Each time the VAE decoder outputs a sequence, we mask everything after the first occurrence of this `(EoS) token`. This way, the VAE should be able to learn that this token represents the end of the sequence. This will help us to generate counterfactuals of varying lengths for a trace of a certain length. If your variational autoencoder is not trained well (i.e. there are still Declare trace constraints violated), the counterfactual algorithm does not work well (empirically tested). Moreover, it is often even impossible to find a counterfactual. The same phenomenon happens when the balance between the \emph{reconstruction loss} and \emph{KL divergence} is not fine-tuned. We use a Long Short-Term Memory (LSTM) neural network as our predictive model. Although our counterfactual generation approach is model-agnostic, we do need a gradient-based predictive model as we perform gradient steps to find the optimal counterfactual(s). For the hyperparameter tuning, we performed a three-fold cross-validation with the use of~\hyperlink{http://hyperopt.github.io/hyperopt/}{hyperopt} for the LSTM neural networks. 

More detailed information about design implementations and parameters for the REVISE$^{+}$ and REVISED$^{+}$ algorithms are provided on the GitHub repository\footnote{\url{https://github.com/AlexanderPaulStevens/Counterfactual-Explanations}}, with the benchmark setting build on top of the work of Pawelczyk et al. \cite{pawelczyk2021carla}.

\subsection{Results}\label{sec: discussion}

The REVISED$^{+}$ approach, when incorporating Declare constraints in the optimisation approach, has a higher average number of counterfactuals per factual (7.6) compared to the approach without constraint optimisation terms (6.28). The approach also generates more plausible counterfactuals on average. This means that adding constraints to the loss functions not only increases the quantity but also enhances the validity \emph{more plausible} of the counterfactuals. 

\begin{table*}[ht]
\centering
\caption{Overview of the results for the REVISE$^{+}$ and the REVISED$^{+}$ counterfactual generation algorithms and the nine different event logs.}
\label{tab: results}
\resizebox{\textwidth}{!}{%
\begin{tabular}{
>{\columncolor[HTML]{FFFFFF}}c |
>{\columncolor[HTML]{FFFFFF}}c 
>{\columncolor[HTML]{FFFFFF}}c |
>{\columncolor[HTML]{FFFFFF}}c |
>{\columncolor[HTML]{FFFFFF}}c 
>{\columncolor[HTML]{FFFFFF}}c |
>{\columncolor[HTML]{FFFFFF}}c 
>{\columncolor[HTML]{FFFFFF}}c 
>{\columncolor[HTML]{FFFFFF}}c |
>{\columncolor[HTML]{FFFFFF}}c 
>{\columncolor[HTML]{FFFFFF}}c 
>{\columncolor[HTML]{FFFFFF}}c |
>{\columncolor[HTML]{FFFFFF}}c |}
\hline
 &
  \cellcolor[HTML]{EFEFEF}Event Log &
  \cellcolor[HTML]{EFEFEF}\begin{tabular}[c]{@{}c@{}}Success\\ Rate $(\%)$\end{tabular} &
  \cellcolor[HTML]{EFEFEF}\begin{tabular}[c]{@{}c@{}}Plausible\\ Rate $(\%)$\end{tabular} &
  \cellcolor[HTML]{EFEFEF}$|CF|$ &
  \cellcolor[HTML]{EFEFEF}y-NN &
  \cellcolor[HTML]{EFEFEF}$L_{1}$ &
  \cellcolor[HTML]{EFEFEF}$L_{2}$ &
  \cellcolor[HTML]{EFEFEF}EMD &
  \cellcolor[HTML]{EFEFEF}$L_{0}$ &
  \cellcolor[HTML]{EFEFEF}DL Edit &
  \cellcolor[HTML]{EFEFEF}LCP &
  \cellcolor[HTML]{EFEFEF}Diversity \\ \cline{2-13} 
\cellcolor[HTML]{FFFFFF} &
  BPIC2012 &
  2.04 &
  {\color[HTML]{CB0000} 0.0} &
  {\color[HTML]{CB0000} \textbf{/}} &
  {\color[HTML]{CB0000} /} &
  {\color[HTML]{CB0000} /} &
  {\color[HTML]{CB0000} /} &
  {\color[HTML]{CB0000} /} &
  {\color[HTML]{CB0000} /} &
  {\color[HTML]{CB0000} /} &
  {\color[HTML]{CB0000} \textbf{/}} &
  {\color[HTML]{CB0000} \textbf{/}} \\
\cellcolor[HTML]{FFFFFF} &
  BPIC2015 (1) &
  8.33 &
  47.95 &
  35.38 &
  0.96 &
  102.17 &
  7.16 &
  60.37 &
  41.07 &
  40.66 &
  3.23 &
  21.63 \\
\cellcolor[HTML]{FFFFFF} &
  BPIC2015 (2) &
  6.68 &
  89.43 &
  40.72 &
  0.79 &
  134.1 &
  8.08 &
  67.1 &
  47.82 &
  47.74 &
  4.79 &
  9.87 \\
\cellcolor[HTML]{FFFFFF} &
  BPIC2015 (3) &
  10.24 &
  62.04 &
  38.63 &
  0.03 &
  102.0 &
  7.0 &
  54.64 &
  36.41 &
  36.09 &
  5.81 &
  16.79 \\
\cellcolor[HTML]{FFFFFF} &
  BPIC2015 (4) &
  8.27 &
  16.24 &
  40.8 &
  0.05 &
  95.85 &
  6.82 &
  36.93 &
  30.41 &
  30.03 &
  4.66 &
  14.4 \\
\cellcolor[HTML]{FFFFFF} &
  BPIC2015 (5) &
  18.28 &
  28.98 &
  45.57 &
  0.34 &
  103.1 &
  6.96 &
  42.83 &
  36.27 &
  36.25 &
  5.97 &
  14.43 \\
\cellcolor[HTML]{FFFFFF} &
  Sepsis (1) &
  1.14 &
  61.23 &
  20.4 &
  0.79 &
  39.24 &
  4.49 &
  1.69 &
  11.57 &
  11.31 &
  6.0 &
  0.88 \\
\cellcolor[HTML]{FFFFFF} &
  Sepsis (2) &
  1.06 &
  93.78 &
  15.3 &
  0.98 &
  32.98 &
  4.21 &
  1.1 &
  10.34 &
  9.88 &
  2.89 &
  0.51 \\
\cellcolor[HTML]{FFFFFF} &
  Sepsis (3) &
  0.5 &
  {\color[HTML]{CB0000} 0.0} &
  {\color[HTML]{CB0000} \textbf{/}} &
  {\color[HTML]{CB0000} /} &
  {\color[HTML]{CB0000} \textbf{/}} &
  {\color[HTML]{CB0000} \textbf{/}} &
  {\color[HTML]{CB0000} \textbf{/}} &
  {\color[HTML]{CB0000} /} &
  {\color[HTML]{CB0000} /} &
  {\color[HTML]{CB0000} /} &
  {\color[HTML]{CB0000} /} \\ \cline{2-13} 
\multirow{-10}{*}{\cellcolor[HTML]{FFFFFF}REVISE+} &
  \cellcolor[HTML]{EFEFEF}(avg.) &
  \cellcolor[HTML]{EFEFEF}6.28 &
  \cellcolor[HTML]{EFEFEF}44.41 &
  \cellcolor[HTML]{EFEFEF}\textbf{26.31} &
  \cellcolor[HTML]{EFEFEF}0.44 &
  \cellcolor[HTML]{EFEFEF}\textbf{67.72} &
  \cellcolor[HTML]{EFEFEF}\textbf{4.97} &
  \cellcolor[HTML]{EFEFEF}\textbf{29.41} &
  \cellcolor[HTML]{EFEFEF}\textbf{23.77} &
  \cellcolor[HTML]{EFEFEF}\textbf{23.55} &
  \cellcolor[HTML]{EFEFEF}3.71 &
  \cellcolor[HTML]{EFEFEF}8.72 \\ \hline
\cellcolor[HTML]{FFFFFF} &
  BPIC2012 &
  2.0 &
  34.93 &
  56.0 &
  0.13 &
  95.84 &
  7.05 &
  13.45 &
  46.0 &
  42.67 &
  4.0 &
  1.05 \\
\cellcolor[HTML]{FFFFFF} &
  BPIC2015 (1) &
  9.36 &
  45.59 &
  35.31 &
  0.93 &
  101.7 &
  7.14 &
  62.83 &
  40.49 &
  40.05 &
  3.35 &
  22.63 \\
\cellcolor[HTML]{FFFFFF} &
  BPIC2015 (2) &
  15.67 &
  72.56 &
  46.68 &
  0.05 &
  134.91 &
  8.11 &
  64.16 &
  47.16 &
  47.09 &
  4.53 &
  13.62 \\
\cellcolor[HTML]{FFFFFF} &
  BPIC2015 (3) &
  10.98 &
  51.23 &
  37.9 &
  0.02 &
  101.34 &
  6.96 &
  54.87 &
  34.86 &
  34.55 &
  6.33 &
  16.07 \\
\cellcolor[HTML]{FFFFFF} &
  BPIC2015 (4) &
  11.33 &
  30.76 &
  40.53 &
  0.8 &
  98.49 &
  6.97 &
  40.94 &
  34.1 &
  33.7 &
  4.58 &
  16.35 \\
\cellcolor[HTML]{FFFFFF} &
  BPIC2015 (5) &
  16.82 &
  27.69 &
  44.87 &
  0.29 &
  102.15 &
  6.91 &
  43.32 &
  37.18 &
  37.12 &
  5.46 &
  17.59 \\
\cellcolor[HTML]{FFFFFF} &
  Sepsis (1) &
  1.43 &
  67.96 &
  20.78 &
  0.79 &
  39.65 &
  4.53 &
  1.51 &
  12.24 &
  12.12 &
  6.01 &
  0.89 \\
\cellcolor[HTML]{FFFFFF} &
  Sepsis (2) &
  0.72 &
  97.96 &
  16.3 &
  0.99 &
  33.03 &
  4.23 &
  1.45 &
  11.05 &
  10.66 &
  3.16 &
  0.55 \\
\cellcolor[HTML]{FFFFFF} &
  Sepsis (3) &
  0.07 &
  4.8 &
  31.61 &
  0.16 &
  54.07 &
  5.47 &
  5.78 &
  24.32 &
  23.99 &
  3.46 &
  0.6 \\ \cline{2-13} 
\multirow{-10}{*}{\cellcolor[HTML]{FFFFFF}REVISED} &
  \cellcolor[HTML]{EFEFEF}(avg.) &
  \cellcolor[HTML]{EFEFEF}\textbf{7.6} &
  \cellcolor[HTML]{EFEFEF}\textbf{55.78} &
  \cellcolor[HTML]{EFEFEF}36.66 &
  \cellcolor[HTML]{EFEFEF}\textbf{0.46} &
  \cellcolor[HTML]{EFEFEF}84.58 &
  \cellcolor[HTML]{EFEFEF}6.37 &
  \cellcolor[HTML]{EFEFEF}32.03 &
  \cellcolor[HTML]{EFEFEF}31.93 &
  \cellcolor[HTML]{EFEFEF}31.33 &
  \cellcolor[HTML]{EFEFEF}\textbf{4.54} &
  \cellcolor[HTML]{EFEFEF}\textbf{9.93} \\ \hline
\end{tabular}%
}
\end{table*}

The results in the remaining columns are based on the existence of plausible counterfactuals. Consequently, for the event logs BPIC2012 and Sepsis (3), no values are available as the plausible rate is zero, which means that the generation algorithm did not produce any plausible counterfactuals for these event logs. For the BPIC2012 (Sepsis (3)), there were 45 (4) trace Declare constraints and 4 (16) label-specific Declare constraints. In this table, we used the 5 nearest neighbours for y-NN. This means that the \emph{y-NN} has a value of 0.8 if 4 out of 5 of the nearest neighbour traces have the same label. Next, the proximity distance metrics $L_{1}$, $L_{2}$ and $EMD$ indicate that by adding the plausible constraints, the counterfactuals (on average) seem to be further away from the factual than without adding the constraints. This means that plausibility comes with a cost of proximity. Second, the algorithm is not optimised for sparsity and therefore the REVISED$^{+}$ algorithm shows relatively high values for $L_{0}$, \emph{DL Edit}, and \emph{LCP}. We argue that we want to obtain insights into how many changes would have been required to have had another prediction, and high sparsity values indicate that many changes would have been requested. Finally, the REVISED$^{+}$ produces more diverse counterfactuals compared to REVISE$^{+}$, which might also be a reason why the metrics for proximity are higher (i.e. more and more diverse counterfactuals lead to counterfactuals with higher proximity values).

This work is therefore the first counterfactual generation algorithm that can generate \emph{feasible} and \emph{plausible} by restricting the control flow behaviour to the process data manifold while adhering to sequential patterns learned with the use of Declare templates. The algorithm REVISED$^{+}$ helps us learn the decision-making process of the predictive model by guiding it through the process data manifold, which is shown in the following case study.

\begin{table*}[]
\centering
\caption{The counterfactuals generated by the REVISED$^+$ generation algorithm for a case of the Sepsis (2) event log.}
\label{tab: onefactual}
\resizebox{\textwidth}{!}{%
\begin{tabular}{|ll|l|ll|lll|lll|l|}
\hline
\rowcolor[HTML]{EFEFEF} 
\multicolumn{1}{|c}{\cellcolor[HTML]{EFEFEF}Case} &
  \multicolumn{1}{c|}{\cellcolor[HTML]{EFEFEF}Counterfactual} &
  \multicolumn{1}{c|}{\cellcolor[HTML]{EFEFEF}Violations $(\%)$} &
  \multicolumn{1}{c}{\cellcolor[HTML]{EFEFEF}$|CF|$} &
  \multicolumn{1}{c|}{\cellcolor[HTML]{EFEFEF}y-NN} &
  \multicolumn{1}{c}{\cellcolor[HTML]{EFEFEF}$L_{1}$} &
  \multicolumn{1}{c}{\cellcolor[HTML]{EFEFEF}$L_{2}$} &
  \multicolumn{1}{c|}{\cellcolor[HTML]{EFEFEF}EMD} &
  \multicolumn{1}{c}{\cellcolor[HTML]{EFEFEF}$L_{0}$} &
  \multicolumn{1}{c}{\cellcolor[HTML]{EFEFEF}DL Edit} &
  \multicolumn{1}{c|}{\cellcolor[HTML]{EFEFEF}LCP} &
  \multicolumn{1}{c|}{\cellcolor[HTML]{EFEFEF}Diversity} \\ \hline
 &
  [3, 5, 4, 9, 2, 8, 7, 6, 1, 9, 9, 2, 2, 9, 9, 16] &
  0.0 &
  16 &
  1.0 &
  28.21 &
  3.66 &
  0.57 &
  10 &
  10 &
  3 &
  0.33 \\
 &
  [3, 5, 4, 9, 2, 2, 7, 6, 1, 9, 9, 2, 2, 9, 9, 160] &
  0.0 &
  16 &
  1.0 &
  28.11 &
  3.65 &
  0.86 &
  10 &
  10 &
  3 &
  0.33 \\
 &
  [3, 5, 4, 9, 2, 2, 7, 6, 1, 9, 9, 2, 2, 9, 9, 2, 16] &
  0.0 &
  17 &
  1.0 &
  27.69 &
  3.61 &
  0.76 &
  10 &
  10 &
  3 &
  0.33 \\
 &
  [3, 5, 4, 9, 2, 2, 7, 6, 1, 9, 9, 2, 2, 9, 9, 9, 16] &
  0.0 &
  17 &
  1.0 &
  27.69 &
  3.61 &
  0.43 &
  9 &
  9 &
  3 &
  0.33 \\
 &
  [3, 5, 4, 9, 2, 2, 7, 6, 1, 9, 9, 2, 2, 9, 9, 9, 2, 16] &
  0.0 &
  18 &
  1.0 &
  27.25 &
  3.57 &
  0.33 &
  8 &
  8 &
  3 &
  0.33 \\
 &
  [3, 5, 4, 9, 2, 2, 7, 6, 1, 9, 9, 2, 2, 9, 9, 9, 2, 2, 16] &
  0.0 &
  19 &
  0.8 &
  26.8 &
  3.53 &
  0.33 &
  7 &
  7 &
  3 &
  0.33 \\
 &
  [3, 5, 4, 9, 2, 2, 7, 6, 1, 9, 9, 2, 2, 2, 9, 9, 2, 2, 16] &
  0.0 &
  19 &
  1.0 &
  26.73 &
  3.52 &
  0.67 &
  8 &
  8 &
  3 &
  0.33 \\
\multirow{-8}{*}{[3, 5, 4, 7, 8, 9, 2, 6, 1, 9, 2, 1, 2, 9, 9, 9, 2, 9, 16]} &
  [3, 5, 4, 9, 2, 2, 7, 6, 1, 9, 9, 2, 2, 2, 9, 9, 9, 2, 16] &
  0.0 &
  19 &
  0.8 &
  26.66 &
  3.51 &
  0.33 &
  7 &
  9 &
  3 &
  0.33 \\ \hline
\end{tabular}%
}
\end{table*}

As a small case study, we use the Sepsis Cases (2) event log, which describes the Sepsis patients with life-threatening symptoms in the Dutch hospital. In the Sepsis Event Log, every case starts with the registration of the patient in the Emergency department (ER), representing a \emph{Trace Declare Constraint} (TDC). Table \ref{tab: TDCLDC constraints} includes \emph{Label-specific Declare Constraints} (LDC), which are sequential patterns only present in the cases where the patient is not admitted to the ICU. For instance, the constraint \textit{succession(ER registration - Admission NC)} implies that ER registration must be followed by admission to the Normal Care (NC) ward. Even stronger, this also indicates that the activity \textit{Admission NC} is also characteristic to cases with a positive prediction, as it is not a Trace Declare Constraint (TDC) applicable to ALL cases.

Both from a humanitarian point of view and a cost-saving perspective of the hospital, we want to avoid that the patient has to be admitted to the ICU (hereinafter referred to as the \emph{negative} prediction). Therefore, we only generate counterfactuals for the patients who are predicted to be admitted to the hospital, and Table \ref{tab: onefactual} represents the results for one case of the event log. The activities in our sequences (in Table \ref{tab: onefactual} are labelled using a label encoding scheme, as follows:
\begin{multicols}{2}
\begin{enumerate}[label=(\arabic*), itemsep=0pt, parsep=0pt, topsep=0pt, partopsep=0pt, left=0pt]
    \item Admission NC
    \item CRP
    \item ER registration
    \item ER sepsis triage
    \item ER triage
    \item IV antibiotics
    \item IV liquid
    \item LacticAcid
    \item Leucocytes
    \item Release A
    \item Release B
    \item Release C
    \item Release D
    \item Return ER
    \item Other
    \item \emph{End of sequence (EoS)}
\end{enumerate}
\end{multicols}

We see, for instance, that the activity \emph{Admission NC} (1) is in all of our plausible counterfactuals, with the activities \textit{CRP} and \textit{Leucocytes} occurring multiple times within the same counterfactual. This might be an indication that the algorithm is hinting towards more frequent measurements of the \textit{CRP} and  \textit{Leucocytes} should have been taken.
\begin{table}[]
\centering
\caption{Trace Declare constraints and label-specific Declare constraints for the sepsis cases where the outcome is either admission to the ICU or not.
}
\label{tab: TDCLDC constraints}
\resizebox{0.5\columnwidth}{!}{%
\begin{tabular}{cl|
>{\columncolor[HTML]{FFFFFF}}c l}
\hline
\multicolumn{2}{c|}{\cellcolor[HTML]{FFFFFF}} &
  \multicolumn{2}{c}{\cellcolor[HTML]{FFFFFF}} \\
\multicolumn{2}{c|}{\cellcolor[HTML]{FFFFFF}} &
  \multicolumn{2}{c}{\cellcolor[HTML]{FFFFFF}} \\
\multicolumn{2}{c|}{\multirow{-3}{*}{\cellcolor[HTML]{FFFFFF}\textbf{Trace Declare Constraints (TDC)}}} &
  \multicolumn{2}{c}{\multirow{-3}{*}{\cellcolor[HTML]{FFFFFF}\textbf{Label-specific Declare Constraints (LDC)}}} \\ \hline
\multicolumn{1}{c|}{1} &
  \cellcolor[HTML]{FFFFFF}start(ER registration) &
  \multicolumn{1}{c|}{\cellcolor[HTML]{FFFFFF}1} &
  \cellcolor[HTML]{FFFFFF}exactly(ER triage) \\
\multicolumn{1}{c|}{2} &
  \cellcolor[HTML]{FFFFFF}exactly(ER registration) &
  \multicolumn{1}{c|}{\cellcolor[HTML]{FFFFFF}2} &
  \cellcolor[HTML]{FFE599}succession(Admission NC - EoS) \\
\multicolumn{1}{c|}{3} &
  \cellcolor[HTML]{FFFFFF}last(EoS) &
  \multicolumn{1}{c|}{\cellcolor[HTML]{FFFFFF}3} &
  \cellcolor[HTML]{FFE599}succession(ER registration - Admission NC) \\
\multicolumn{1}{c|}{4} &
  \cellcolor[HTML]{FFFFFF}exactly(Eos) &
  \multicolumn{1}{c|}{\cellcolor[HTML]{FFFFFF}4} &
  \cellcolor[HTML]{FFFFFF}succession(ER registration - ER triage) \\
\multicolumn{1}{c|}{5} &
  \cellcolor[HTML]{FFFFFF}succession(ER registration - EoS) &
  \multicolumn{1}{c|}{\cellcolor[HTML]{FFFFFF}5} &
  \cellcolor[HTML]{FFFFFF}succession(ER triage - EoS) \\
\multicolumn{1}{c|}{6} &
  \cellcolor[HTML]{FFFFFF}not succession(EoS - ER registration) &
  \multicolumn{1}{c|}{\cellcolor[HTML]{FFFFFF}6} &
  \cellcolor[HTML]{FFE599}not succesion(Admission NC - ER registration) \\ \cline{1-2}
 &
  \multicolumn{1}{c|}{} &
  \multicolumn{1}{c|}{\cellcolor[HTML]{FFFFFF}7} &
  \cellcolor[HTML]{FFFFFF}not succesion(ER triage - ER registration) \\
 &
  \multicolumn{1}{c|}{} &
  \multicolumn{1}{c|}{\cellcolor[HTML]{FFFFFF}8} &
  \cellcolor[HTML]{FFE599}not succession(EoS - Admission NC) \\
 &
  \multicolumn{1}{c|}{} &
  \multicolumn{1}{c|}{\cellcolor[HTML]{FFFFFF}9} &
  \cellcolor[HTML]{FFFFFF}not succession(EoS  - ER triage) \\ \cline{3-4} 
\end{tabular}%
}
\end{table}
\section{Related Work}\label{sec: relatedwork}

In this related work section, we discuss and compare the most important and relevant counterfactual algorithms in the field of XAI, and complement this with the existing counterfactual generation algorithms in the field of predictive process analytics.  

\subsection{Counterfactual algorithms in the field of XAI}

For a more exhaustive overview of the different counterfactual algorithms in the field of XAI, we refer to the benchmark studies of Chou et al. \cite{chou2022counterfactuals} and Verma et al. \cite{verma2020counterfactual}. In this subsection, we mostly focus on the subset of the counterfactual generation algorithm implemented in the CARLA benchmark \cite{pawelczyk2021carla}. 
 
As established in Section \ref{subsec: counterfactualsPPA}, our emphasis lies on approaches that optimise for both plausibility and feasibility, the latter being achieved through a manifold-based methodology, as previous work already demonstrated that out-of-distribution examples have an impact on the learning abilities of predictive models \cite{stevens2022assessing, stevens2023manifold}.

One of the pioneers concerning the counterfactual generation algorithm comes from Wachter et al. \cite{wachter2017counterfactual}. The approach focuses on minimising the property \emph{proximity}, however, it does not explicitly consider factors such as \emph{plausibility} or \emph{feasibility}. The DICE algorithm \cite{mothilal2020explaining} is an extension of the algorithm of Wachter et al., and additionally focuses on \emph{diversity} by generating multiple, diverse counterfactuals to allow the user to choose the counterfactual they deem more interpretable or understandable. Another established algorithm is the Growing Spheres algorithm \cite{laugel2018comparison}, which is specifically optimised for both \emph{proximity} and \emph{sparsity}. Next, The FOCUS algorithm \cite{lucic2022focus} is tailored for \emph{proximity}. In contrast, the FeatureTweak algorithm \cite{tolomei2017interpretable} optimises for both \emph{proximity} and \emph{plausibility} by distinguishing between mutable and immutable features. However, this model-agnostic approach does not consider \emph{feasibility} as it does not take into account the distribution of the data (or the correlation between features). Finally, the paper Actionable Recourse \cite{ustun2019actionable} tackles \emph{feasibility} through \emph{actionability} by constraining the generated counterfactuals to not alter immutable features nor alter mutable features in an \emph{infeasible way}, but by not taking into account the data manifold,  possible correlation effects/interdependencies between features are disregarded.
 
The FACE counterfactual algorithm \cite{poyiadzi2020face} is an example-based counterfactual algorithm that states to generate feasible and actionable counterfactual explanations. As the generated counterfactuals are existing examples from the dataset (i.e. \emph{example-based} explanations), the counterfactuals are always \emph{feasible}. Furthermore, the authors define the actionability of this approach by implying that the generation approach must follow a high-density path. Although this is a very interesting and valuable approach, the use of an example-based counterfactual explanation algorithm means that the counterfactual algorithm is dependent on the quality (and quantity) of your dataset. By using existing instances as a counterfactual explanation, we argue this might induce some problems in the field of PPA for event logs with a lot of (trace) variants, i.e. every trace is almost unique. Another drawback lies in the fact that it is hard to find a quantitative measure for the distance between traces (as two traces are connected with edges if they are deemed \emph{similar} enough). Moreover, in our approach, we assume that counterfactuals can have a different length than the existing query we are looking for a counterfactual for. This would have to be resembled in this distance metric. A final limitation, as highlighted by the authors themselves, is that the algorithm is dependent on the $\epsilon$, which determines the edges between instances. The REVISE counterfactual algorithm \cite{joshi2019towards} uses variational autoencoders to learn the distribution of the data, and is an iterative approach as it performs gradient steps in the latent space to generate counterfactuals. Similarly, the CCHVAE \cite{pawelczyk2020learning} approach generates counterfactuals by sampling points from the latent space of a VAE trained on the data and then passing the points through the decoder. Finally, some works use counterfactual explanations for other applications. The CRUDS algorithm \cite{downs2020cruds} extends REVISE with the causal constraints from Karimi et al. \cite{karimi2021algorithmic} as a means to go from counterfactual explanations to interventions. Next, the CLUE \cite{antoran2020getting} algorithm uses counterfactual explanations to improve the confidence of a classifier in its prediction. More specifically, Bayesian neural networks are used to generate uncertainty estimates of their predictions to find the smallest change that could be made to an input, while keeping it in distribution. The Contrastive Explanations Method (CEM) \cite{dhurandhar2018explanations} also trains a Variational Autoencoder (VAE) on the data distribution. Its objective is to identify the minimal components required in the input to justify its prediction, together with what should be minimally and necessarily absent while still being able to distinguish the input from another closely related input but with another predicted label.

Although there already exist of lot of interesting approaches, most of the current generation algorithms rely on domain knowledge or are invalid/infeasible \cite{chou2022counterfactuals}. Furthermore, generation algorithms that solely optimise on proximity loss and not inherently optimise for plausibility might generate infeasible or impractical solutions \cite{moreira2022benchmarking}. 

\subsection{Counterfactuals in the field of PPA}\label{subsec: counterfactualsPPA}

Counterfactuals help to reason about how different activities influence the outcome of process execution and what has to change in the execution to change the predicted outcome \cite{buliga2023counterfactuals}

The CREATED algorithm, as presented in Hundogan et al. \cite{hundogan2023created}, is an extension of the employed evolutionary models for the generation of counterfactuals, prioritising feasibility, similarity, and sparsity. Feasibility is assessed using Markov models, while similarity is gauged through a weighted adaptation of the Damerau-Levenstein distance, and sparsity is quantified as the count of differences between event attributes. Unlike manifold-based approaches, evolutionary algorithms utilised by CREATED are not confined to generating counterfactuals within the bounds of the learned data distribution. 
In this approach, process constraints are not specifically taken into account, which means that the plausibility of the generated counterfactual is therefore dependent on the quality of their Markov model (i.e. are there any process constraints violated?). The algorithm can generate diverse counterfactuals and does not require any domain knowledge. On the other hand, LORELEY, an extension of the LORE technique \cite{guidotti2019factual} introduced by Huang et al. \cite{huang2021counterfactual}, addresses control flow attribute representation challenges. It denotes all the control flow attributes as a single attribute to avoid issues related to control flow constraints and domain knowledge during crossover and mutation steps. This generative approach can produce multiple, diverse counterfactuals and therefore is optimised for diversity.

Next, several counterfactual generation approaches have already been introduced to the field of next activity prediction, another subfield of PPA that is closely related to process outcome prediction. DICE4EL \cite{hsieh2021dice4el} is a milestone-aware counterfactual generation algorithm that helps to reason about what would have to be changed to a loan application in order to meet the desired milestone (i.e. why is the predicted next activity different from the desired milestone?). It uses these milestone activities to select the counterfactual candidates from the original dataspace and uses a user-defined loss function to update these candidate counterfactuals using gradient descent. Therefore, this algorithm considers properties such as proximity and feasibility (through the defined scenario loss).

In the field of PPA, counterfactual explanations are also used for other applications. In \cite{narendra2019counterfactual}, structural causal models are used for counterfactual reasoning about how the process performance indicators can be improved. This approach used BPM models and event logs simulated from these BPM models for causal structure learning and answering intervention questions.

Buliga et al. \cite{buliga2023counterfactuals} introduced an evaluation framework for counterfactual generation algorithms in PPA as a means to define what a good counterfactual explanation for a trace is and to introduce evaluation metrics to evaluate the quality of a counterfactual in PPA tasks. The current state-of-the-art evaluation metrics for counterfactual approaches to the PPA domain are adapted by introducing a new metric that measures the compliance of the returned counterfactual to specific process constraints (i.e. the SAT score). In addition, the authors also state that there is a lack of consolidated methods for generating the desired good counterfactual explanations. 

\section{Conclusion}\label{sec: conclusion}

In this paper, we present an innovative approach dedicated to deriving plausible counterfactual explanations specifically tailored for process outcome predictions. To ensure the practical utility and efficacy of our generated counterfactuals, we constrain our counterfactual algorithm to produce counterfactuals that lie within a high-density region and therefore assure their real-world applicability. Moreover, we augment our variational autoencoder with a set of essential constraints by incorporating \emph{process trace} constraints into our loss function. Additionally, \emph{label-specific process constraints} are integrated into our counterfactual generation algorithm. These constraints add an extra layer of precision to the counterfactual generation process. Both of these constraints are mined from Declare patterns, which is often used to describe behavioural characteristics in the field of predictive process analytics.

In the pursuit of a holistic evaluation, we introduce a post-hoc evaluation framework. This framework assesses crucial properties such as \emph{proximity}, \emph{sparsity},\emph{plausibility},\emph{feasibility} and \emph{diversity}.
The conclusion of this paper is that counterfactual explanation methods optimized for plausibility and feasibility are an important step towards more trustworthy explanations for process outcome prediction tasks. The proposed method has the potential to improve decision-making in various domains by providing more interpretable explanations for complex models.

\section{Acknowledgements}
This study was financed by the Research Foundation Flanders under
grant number V440623N as well as grant number G039923N, by KU Leuven, Belgium under project 3H200414, and Internal Funds KU Leuven under grant number C14/23/031. 

\bibliographystyle{unsrt}  
\bibliography{references}

\end{document}